\documentclass[conference]{IEEEtran}
\IEEEoverridecommandlockouts
\usepackage[backend=biber]{biblatex}
\usepackage{amsmath,amssymb,amsfonts}
\usepackage{algorithmic}
\usepackage{graphicx, subcaption}
\usepackage{textcomp}
\usepackage{xcolor}
\usepackage{hyperref}
\def\BibTeX{{\rm B\kern-.05em{\sc i\kern-.025em b}\kern-.08em
    T\kern-.1667em\lower.7ex\hbox{E}\kern-.125emX}}
\begin{document}

\title{RuCLIP - new models and experiments: \\ a technical report}

\author{\IEEEauthorblockN{Alex Shonenkov\IEEEauthorrefmark{1},
Andrey Kuznetsov\IEEEauthorrefmark{1}\IEEEauthorrefmark{3}, Denis Dimitrov\IEEEauthorrefmark{1}\IEEEauthorrefmark{4}, Tatyana Shavrina\IEEEauthorrefmark{2},
Daniil Chesakov\IEEEauthorrefmark{1}, \\ 
Anastasia Maltseva\IEEEauthorrefmark{1},
Alena Fenogenova\IEEEauthorrefmark{1}, Igor Pavlov\IEEEauthorrefmark{1}, Anton Emelyanov\IEEEauthorrefmark{1}, Sergey Markov\IEEEauthorrefmark{2}, \\ Daria Bakshandaeva\IEEEauthorrefmark{1}, Vera Shybaeva\IEEEauthorrefmark{1} and Andrey Chertok\IEEEauthorrefmark{1}}

\IEEEauthorblockA{
\IEEEauthorrefmark{1} Sber AI,
\IEEEauthorrefmark{2} Sber Devices,
\IEEEauthorrefmark{3} Samara National Research University,
\IEEEauthorrefmark{4} Lomonosov Moscow State University
\\
Email: \{AVShonenkov, AVladimirKuznetsov, Dimitrov.D.V\}@sberbank.ru,}
}

\maketitle

\begin{abstract}
In the report we propose six new implementations of ruCLIP model trained on our 240M pairs. The accuracy results are compared with original CLIP model with Ru-En translation (OPUS-MT) on 16 datasets from different domains. Our best implementations outperform CLIP + OPUS-MT solution on most of the datasets in few-show and zero-shot tasks. In the report we briefly describe the implementations and concentrate on the conducted experiments. Inference execution time comparison is also presented in the report.

\end{abstract}


\section{Introduction}
The problem of the lack of expensive labeled data for training new deep architectures is becoming more and more acute every year. Due to the lack of resources to perform large-scale data collection and labeling, scientists in the modern world are moving towards the development of self-supervised learning solutions that will allow them to receive information about new objects, focusing on previously acquired knowledge.

Another problem is that deep learning models are often developed for a specific domain (and one modality - text, code, sound, image or time series). At the same time, as practice shows, human learning is multimodal in nature (and teaching a person to solve a problem in one modality helps the process of teaching the same person to solve some other problem in another modality). In this regard, we assume that in the case of deep learning models, such an effect can also take place: models working in specific modalities can extract missing knowledge from other modalities or even when solving some specific problems.

In 2021, language and visual transformers were especially actively developed, which led to the emergence of new developments in the field of multimodal data analysis. A striking example was the DALL-E\cite{dalle} model for image synthesis from text descriptions from OpenAI. We also successfully broke into this race of transformers and could not ignore another useful application of language and visual transformers, the CLIP (Contrastive Language - Image Pre-training)\cite{clip} model from OpenAI, whose task is to determine the 'semantic' proximity of texts in natural (English) language and images. Successful training of the model by the OpenAI team made it possible to solve such computer vision problems as zero-shot classification and zero-shot object detection. As a result, we set ourselves the goal of CLIP model Russification so that the model can be used directly in Russian and bypass the use of translators in conjunction with the original English CLIP model. We made the first steps in this direction at the beginning of 2021 - we developed the ruCLIP Small model. Now we want to describe six more versions of the ruCLIP model that we have trained during this time.

\section{RuCLIP Models}
 We releases the following versions of the ruCLIP model, which differ in the number of layers of the ViT encoder, the size of the patch used (14×14, 16×16, 32×32) and the sizes of input images — 224×224, 336×336 and 384×384 pixels. The semantics of the model name is as follows:
 \begin{itemize}
     \item ruCLIP Base [vit-base-patch16-224];
     \item ruCLIP Base [vit-base-patch32-224];
     \item ruCLIP Base [vit-base-patch32-384];
     \item ruCLIP Large [vit-large-patch14-224];
     \item ruCLIP Base [vit-base-patch16-384];
     \item ruCLIP Large [vit-large-patch14-336].
 \end{itemize}
 
 All new versions of the ruCLIP model were trained on open datasets, as well as on data from the Sber ecosystem. In total, we managed to collect about 240 million unique 'image-description' pairs. We trained the model for 12 days on the SberCloud ML Space platform and the Christofari Neo supercomputer using distributed training on 256 Tesla A100 GPUs, which significantly exceeds the resources spent during the first ruCLIP Small training. It is important to emphasize that in the new versions, the batch size 32768 was used during training, which corresponds to how the original CLIP model was trained (in the ruCLIP Small version, the batch size was 16).

The new versions of the ruCLIP model are still based on two components:
\begin{enumerate}
    \item Image Encoder is an encoder that generates a vector representation of images. The well-known ViT remains at the heart of our model.
    \item Text Encoder is an encoder that generates a vector representation of text descriptions. Unlike the ruCLIP Small model, we did not use RuGPT3Small, but took a text transformer with the following parameters:
    \begin{itemize}
        \item Base versions:
        \begin{itemize}
            \item Context Length: 77
            \item Transformer Layers: 12
            \item Transformer Width: 512
            \item Transformer Heads: 8
        \end{itemize}
        \item Large versions:
        \begin{itemize}
            \item Context Length: 77
            \item Transformer Layers: 12
            \item Transformer Width: 768
            \item Transformer Heads: 12
        \end{itemize}
    \end{itemize}
\end{enumerate}

\section{Training Procedure}
\subsection{Data}
As we mentioned earlier to train the ruCLIP models we used a specifically collected dataset, consisting of 240M pairs of 'image-description' pairs. It should be noted that our dataset is one of the largest Russian-language datasets containing 'image-description' pairs, but in comparison to the existing English-language datasets our set is far from the leaders of the list in terms of volume, which can be seen in Table~\ref{tab1}.

\begin{table}[htbp]
\caption{Dataset size comparison}
\begin{center}
\begin{tabular}{|l|c|c|c|c|c|}
\hline
\textbf{Model} & ruCLIP Small & ruCLIP & CLIP & ALIGN & BASIC \\
\hline
\textbf{\shortstack{Num. of \\ samples}} & 3M & 240M & 400M & 1.8B & 6.6B \\
\hline
\end{tabular}
\label{tab1}
\end{center}
\end{table}

The data volume leader belongs to Google with their ALIGN\cite{align} and BASIC\cite{basic} models. In the first of them, data is collected on the basis of Conceptual Captions with simplified data filtering. In the second model, the dataset for ALIGN is supplemented with a private JFT-3B dataset.

\subsection{Training Details}
The training process was quite challenging, so we provide some details in this section. First, we started with a model based on ViT-Large. At the first stage, ruCLIP Large [vit-large-patch14-224] was trained from scratch based on the collected dataset. This model was trained for 380K iterations. Further, by changing the size of the input data to $336\times336$ pixels and training for another 15K iterations, the ruCLIP Large [vit-large-patch14-336] exclusive model was obtained. As for the ViT-Base versions of the model, they were trained by fine-tuning the English model for 140K iterations. As a result, the ruCLIP Base [vit-base-patch16-224] and ruCLIP Base [vit-base-patch32-224] models were obtained. By changing the size of the input images to $384\times384$ pixels and performing fine-tuning for another 20K iterations, the ruCLIP Base [vit-base-patch32-384] and ruCLIP Base [vit-base-patch16-384] exclusive models were obtained.

The training process is visualized below in Fig.~\ref{fig1} as a dependency of the loss value from the number of train iterations. Also, for each training process we set a specific learning rate scheme (right column in Fig.~\ref{fig1}).

\begin{figure}[htbp]
\centerline{\includegraphics[width=0.5\textwidth]{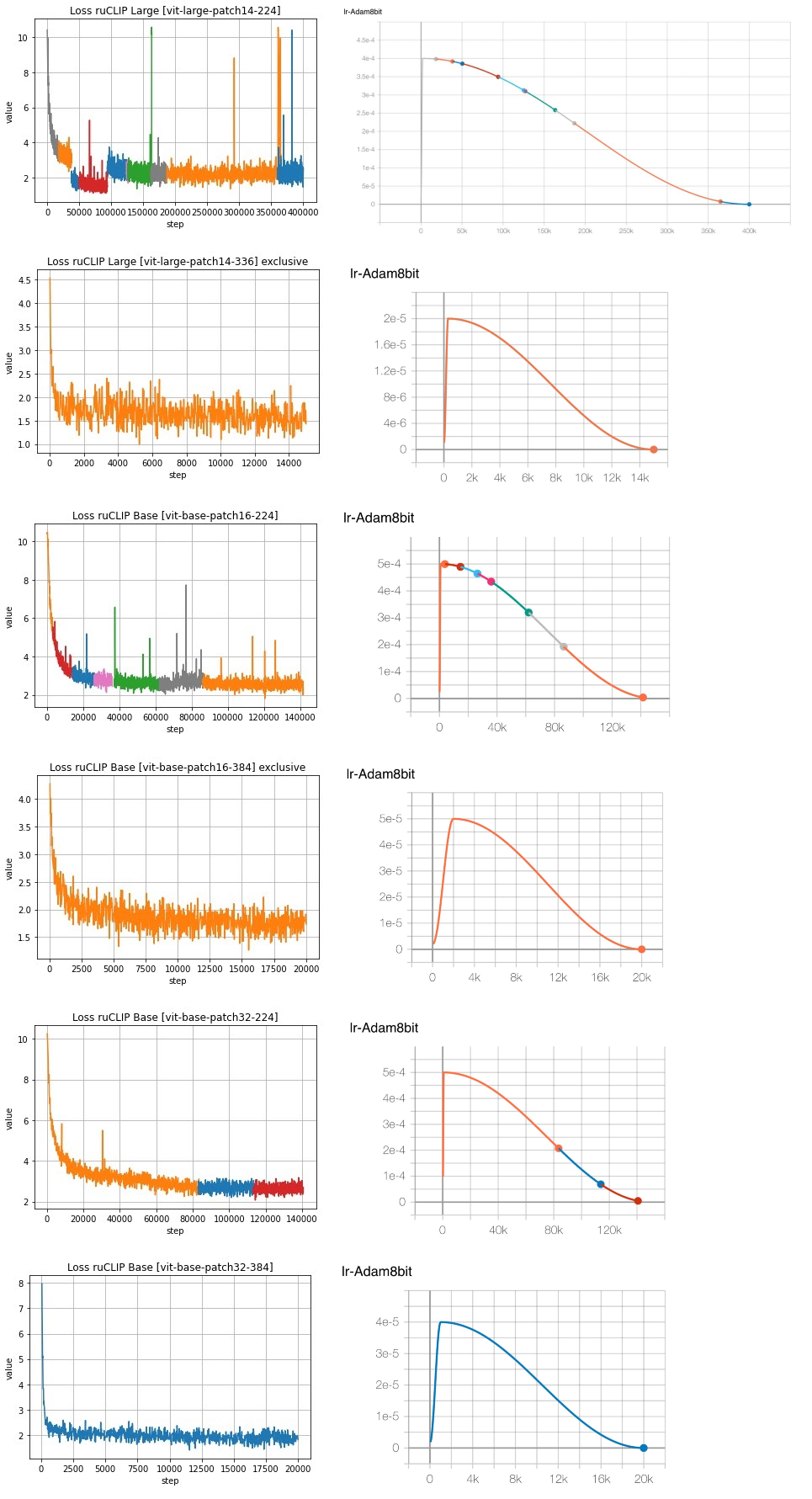}}
\caption{Training process results for ruCLIP architectures.}
\label{fig1}
\end{figure}

\section{Experiments}
During the experiments, we pursued several goals. The first is to see how good different ruCLIP models are on data of different nature, and the second is to compare the quality of our solution with a combination of the OPUS-MT Ru-En translator and the original CLIP model. In the following part of the section we provide the results of our research.

\subsection{Quality Estimation}
The quality of the model is assessed using the standard cosine metric. Text and image descriptions are transformed into their separate embeddings, then we calculate the similarity and choose the best "image-description" pair.

\begin{figure}[htbp]
\centerline{\includegraphics[width=0.5\textwidth]{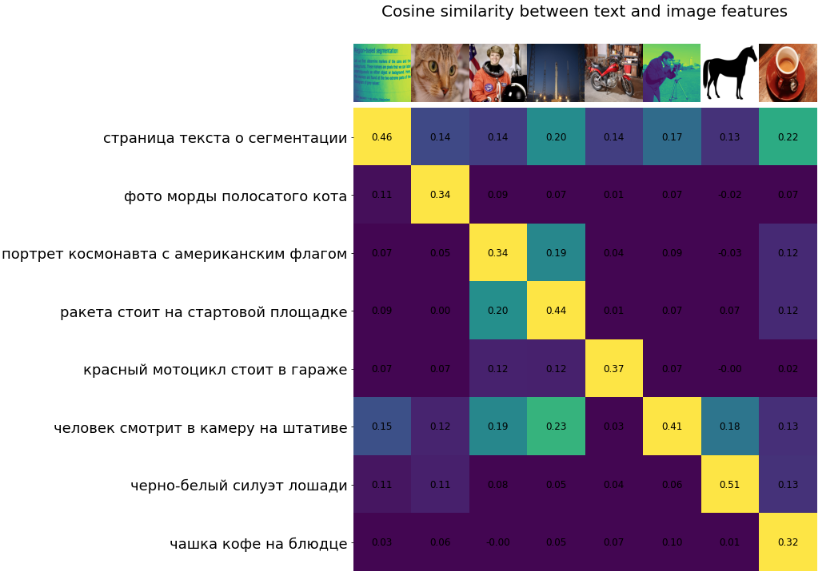}}
\caption{Image and text cosine similarity comparison.}
\label{fig2}
\end{figure}

Further we take the largest of our open models ruCLIP Large [vit-large-patch14-224] and check its quality on a number of "image-description" pairs used for evaluation in the original CLIP\cite{clip} paper. The cosine similarity between text prompts and images for the ruCLIP Large model [vit-large-patch14-224] is shown in Fig.~\ref{fig2}.

In Table~\ref{tab2} we compare the implementations of the ruCLIP model within the zero-shot classification accuracy for different domains datasets. The table shows the results of comparing the first trained ruCLIP Small model, all 6 implementations of the ruCLIP model, the CLIP original model with the Ru-En OPUS-MT translator, and the CLIP original model. In 8 out of 18 datasets, the ruCLIP Large [vit-large-patch14-336] exclusive model shows the best result among all compared, and in those where it loses, the difference is not significant (perhaps, except for MNIST and FGVC Aircraft - there is still something to work on hard). It is worth noting that on the MNIST dataset, even the largest ALIGN and BASIC models cannot outperform CLIP in terms of quality. For most datasets (11 out of 18), the models we trained outperform the CLIP [vit-base-patch16-224] original + OPUS-MT model. The advantage of the CLIP original model (right column) can be explained by the two times larger training sample size (400M vs 240M).

\begin{table*}[t]
\caption{Zero-shot classification accuracy for different models}
\begin{center}
\begin{tabular}{|p{1.5cm}|p{1.5cm}|p{1.5cm}|p{1.5cm}|p{1.5cm}|p{1.5cm}|p{1.5cm}|p{1.5cm}|p{1.5cm}|p{1.5cm}|}
\hline
& \multicolumn{1}{|p{1.5cm}|}{\centering ruCLIP Small \\ \textit{rugpt3-small}}
& \multicolumn{1}{|p{1.5cm}|}{\centering ruCLIP Base \\ \textit{vit-base-patch32-224}}
& \multicolumn{1}{|p{1.5cm}|}{\centering ruCLIP Base \\ \textit{vit-base-patch16-224}}
& \multicolumn{1}{|p{1.5cm}|}{\centering ruCLIP Large \\ \textit{vit-large-patch14-224}}
& \multicolumn{1}{|p{1.5cm}|}{\centering ruCLIP Base \\ \textit{vit-base-patch32-384}}
& \multicolumn{1}{|p{1.5cm}|}{\centering ruCLIP Large \\ \textit{vit-large-patch14-336 \\ exclusive}}
& \multicolumn{1}{|p{1.5cm}|}{\centering ruCLIP Base \\ \textit{vit-base-patch16-384 \\ exclusive}}
& \multicolumn{1}{|p{1.5cm}|}{\centering CLIP \\ \textit{vit-base-patch16-224} \\ + OPUS-MT}
& \multicolumn{1}{|p{1.5cm}|}{\centering CLIP \\ \textit{vit-base-patch16-224} }\\
\hline

Food101, acc & 0.137 & 0.505 & 0.552 & 0.597 & 0.642 & \textbf{0.712} & 0.689 & 0.664 & 0.883 \\\hline

CIFAR10, acc & 0.808 & 0.818 & 0.810 & 0.878 & 0.862 & \textbf{0.906} & 0.845 & 0.859 & 0.893 \\\hline

CIFAR100, acc & 0.440 & 0.504 & 0.496 & 0.511 & 0.529 & 0.591 & 0.569 & \textbf{0.603} & 0.647 \\\hline

Birdsnap, acc & 0.036 & 0.115 & 0.117 & 0.172 & 0.161 & \textbf{0.213} & 0.195 & 0.126 & 0.396 \\\hline

SUN397, acc & 0.036 & 0.452 & 0.462 & 0.484 & 0.510 & 0.523 & 0.521 & 0.447 & 0.631 \\\hline

Stanford Cars, acc & 0.023 & 0.433 & 0.487 & 0.559 & 0.572 & \textbf{0.659} & 0.626 & 0.567 & 0.638 \\\hline

DTD, acc & 0.169 & 0.380 & 0.401 & 0.370 & 0.390 & 0.408 & \textbf{0.421} & 0.243 & 0.432 \\\hline

MNIST, acc & 0.137 & 0.447 & 0.464 & 0.337 & 0.404 & 0.242 & 0.478 & \textbf{0.559} & 0.559 \\\hline

STL10, acc & 0.910 & 0.932 & 0.932 & 0.934 & 0.946 & 0.956 & 0.964 & 0.967 & \textbf{0.970} \\\hline

PCam, acc & 0.484 & 0.501 & 0.505 & 0.520 & 0.506 & 0.554 & 0.501 & \textbf{0.603} & 0.573 \\\hline

CLEVR, acc & 0.104 & 0.148 & 0.128 & 0.152 & 0.188 & 0.142 & 0.132 & \textbf{0.240} & 0.240 \\\hline

Rendered SST2, acc & 0.483 & 0.489 & 0.527 & 0.529 & 0.508 & \textbf{0.539} & 0.525 & 0.484 & 0.484 \\\hline

ImageNet, acc & -- & 0.375 & 0.401 & 0.426 & 0.451 & \textbf{0.488} & 0.482 & 0.392 & 0.638 \\\hline

FGVC Aircraft, mean-per-class & 0.020 & 0.033 & 0.043 & 0.046 & 0.053 & 0.075 & 0.046 & \textbf{0.220} & 0.244 \\\hline

Oxford Pets, mean-per-class & 0.462 & 0.560 & 0.595 & 0.604 & 0.587 & 0.546 & \textbf{0.635}  & 0.507 & 0.874 \\\hline

Caltech101, mean-per-class & 0.590 & 0.786 & 0.775 & 0.777 & 0.834 & \textbf{0.835} & \textbf{0.835} & 0.792 & 0.883 \\\hline

Flowers102, mean-per-class & 0.063 & 0.401 & 0.388 & 0.455 & 0.449 & \textbf{0.517} & 0.452 & 0.357 & 0.697 \\\hline

Hateful Memes, roc-auc & 0.527 & 0.564 & 0.516 & 0.530 & 0.537 & 0.519 & 0.543 & \textbf{0.579} & 0.589 \\

\hline
\end{tabular}
\label{tab2}
\end{center}
\end{table*}

\begin{figure}[htbp]
\centerline{\includegraphics[width=0.5\textwidth]{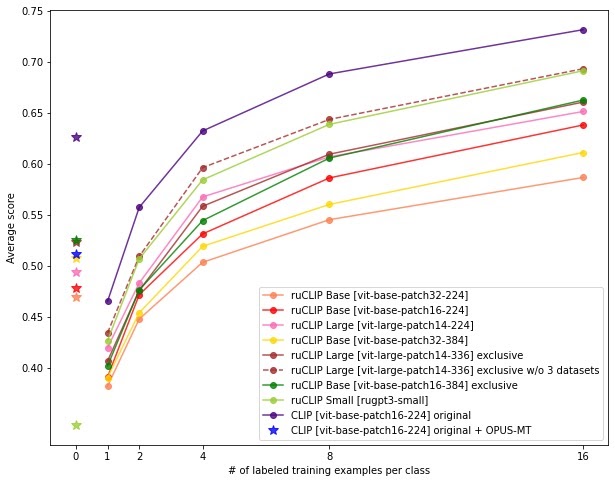}}
\caption{Average the results of zero-shot (asterisks) and few-shot (lines)  experiments for various models for all datasets used in the quality assessment.}
\label{fig3}
\end{figure}

\begin{figure}[htbp]
\centerline{\includegraphics[width=0.5\textwidth]{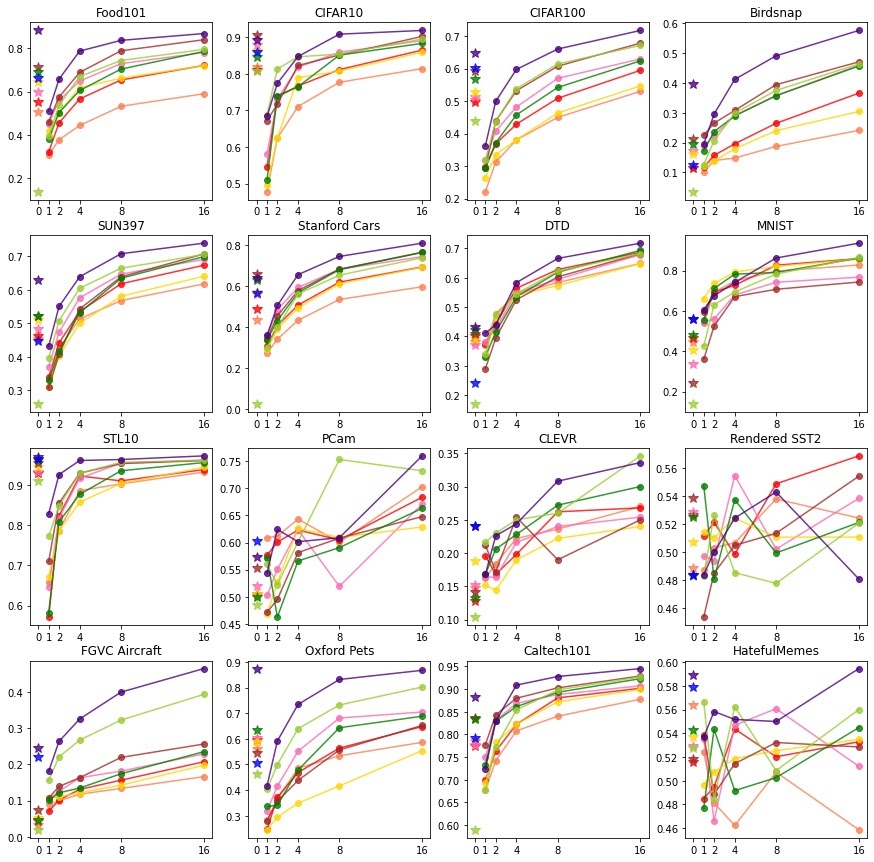}}
\caption{The results of the zero-shot (asterisks) and few-shot (lines) experiments for various models for each of the datasets used in the quality assessment.}
\label{fig4}
\end{figure}

In Fig.~\ref{fig3} and Fig.~\ref{fig4} we show the results of conducted experiments on zero-shot and few-shot classification by 10 models on 16 datasets. Just like in the original article, linear classifiers were trained on the features that CLIP extracts for images using 1, 2, 4, 8, and 16 images for each class. This is a fairly simple and understandable transfer learning approach, called a linear probe, in which the trained model is used to extract features, and then the "head" is trained on the required number of classes with a limited set of images for each class. Since the features that are extracted from the CLIP original and CLIP original + OPUS-MT models are the same, there is no separate few-shot classification plot for CLIP original + OPUS-MT. We also calculated the average few-shot graph for the best ruCLIP Large exclusive model without taking into account three datasets - PCam, Oxford Pets and FGVC Aircraft, on which the model loses significantly, and you can see (dotted line) that the average quality even slightly exceeds ruCLIP Small. The asterisks show the average zero-shot accuracy values for the models.

It should be mentioned that when implementing ruCLIP Small, an Image Encoder with frozen weights of the CLIP original model was used as the backbone and two additional linear layers were added and trained. Due to the fact that only the visual component of the model was used for the linear probe experiment, the quality of ruCLIP Small expected to be very close to CLIP original. This is why ruCLIP Small scores are high and outperform several architectures.

\subsection{Inference Time Estimation}
We did not limit experiments to the qualitative only, but also evaluated the performance of our ruCLIP model implementations on the CIFAR100 dataset using Nvidia-V100 GPU. The results of inference time are shown in Table~\ref{tab3}.

\begin{table}[htbp]
\caption{Inference time (iter/sec) of the ruCLIP model implementations on the CIFAR100 dataset}
\begin{center}
\begin{tabular}{|c|c|c|c|c|c|}
\hline
\multicolumn{1}{|p{1cm}|}{\centering ruCLIP Base \\ vit-b-patch32-224}
& \multicolumn{1}{|p{1cm}|}{\centering ruCLIP Base \\ vit-b-patch16-224}
& \multicolumn{1}{|p{1cm}|}{\centering ruCLIP Large \\ vit-large-patch14-224}
& \multicolumn{1}{|p{1cm}|}{\centering ruCLIP Base \\ vit-base-patch32-384}
& \multicolumn{1}{|p{1cm}|}{\centering ruCLIP Large \\ vit-large-patch14-336 \\ exclusive}
& \multicolumn{1}{|p{1cm}|}{\centering ruCLIP Base \\ vit-base-patch16-384 \\ exclusive} \\
\hline
308.84 & 155.35 & 49.95 & 147.26 & 22.11 & 61.79 \\
\hline
\end{tabular}
\label{tab3}
\end{center}
\end{table}

\section{Conclusion}
\begin{figure}[ht]
    \centering
    
    \begin{subfigure}{\linewidth}
    \centering
    \includegraphics[width=.8\textwidth]{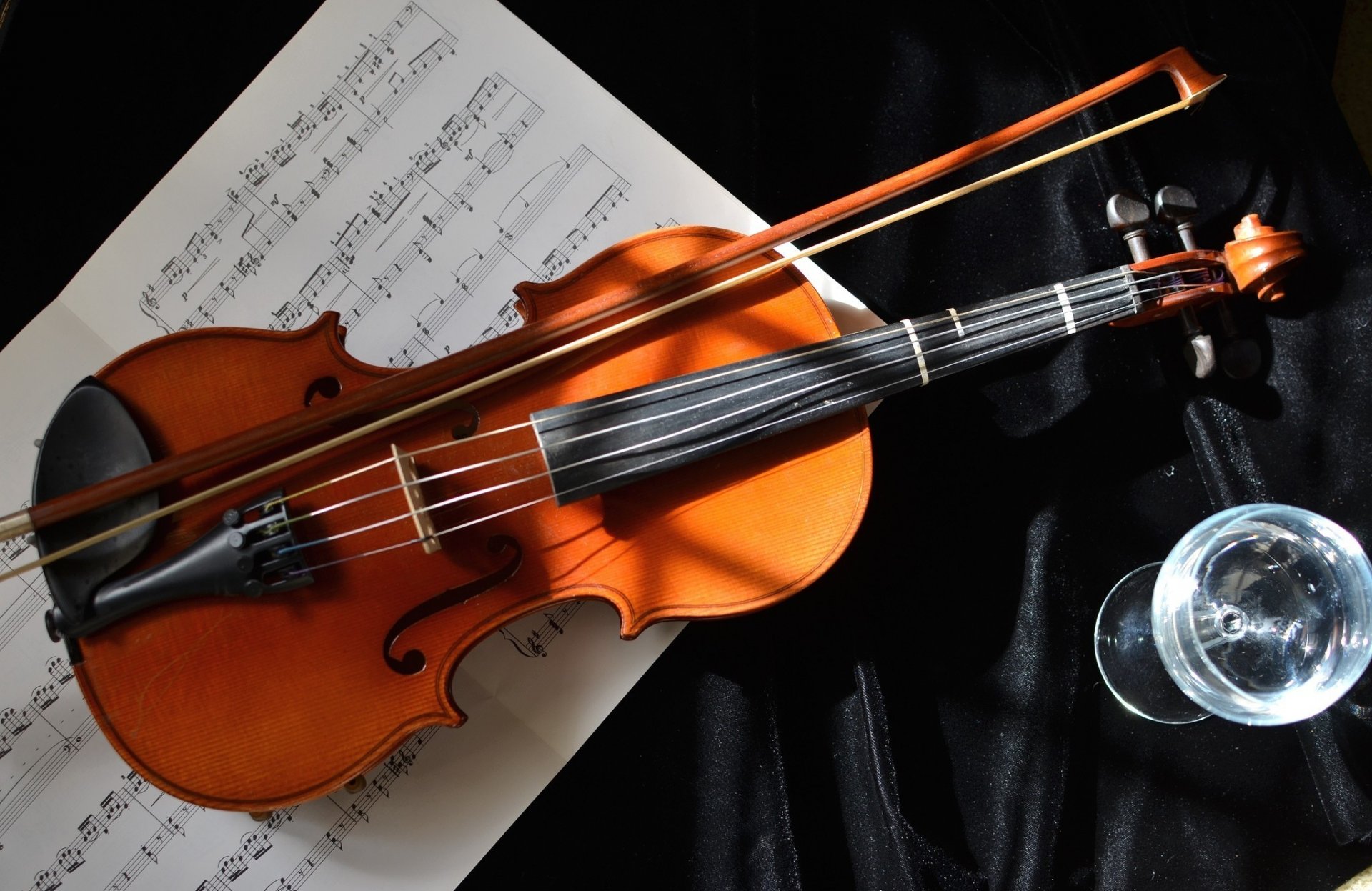}
    \end{subfigure}
    
    \begin{subfigure}{\linewidth}
    \centering
    \includegraphics[width=.5\textwidth]{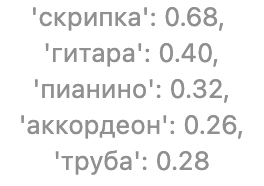}
    \end{subfigure}
    
    \caption{RuCLIP model evaluation example.}
    \label{fig5}
\end{figure}

We managed to train several different versions of ruCLIP, which on a number of datasets successfully outperformed the original CLIP model with a Ru-Eng translator. All the training was based on our dataset of 240M pairs and took 12 days on 256 Tesla GPU A100 of the Christofari Neo supercomputer. Extensive research on datasets from various domains shown the applicability of the ruCLIP model in zero-shot and few-shot classification problems. An example of the ruCLIP model is presented in Fig.~\ref{fig5}).

In the zero-shot and few-shot classification tasks, the ruCLIP model showed lower quality in comparison to the original CLIP model due to the fact that a smaller dataset was used for training our ruCLIP implementations. It should also be noted that the quality of the models increases with the increase in the volume of training samples, which is confirmed by the advantage of ALIGN and BASIC over CLIP due to the significantly larger amount of data used.

\printbibliography

\end{document}